\documentclass{article}
\usepackage[multiple]{footmisc}
\usepackage[nonatbib]{neurips_2022}
\nolinenumbers

\usepackage[bookmarksnumbered,pdfpagelabels=true,plainpages=false,colorlinks=true,
            linkcolor=red,citecolor=red,urlcolor=blue]{hyperref}
            
\usepackage[utf8]{inputenc} 
\usepackage[T1]{fontenc}    
\usepackage{hyperref}       
\usepackage{url}            
\usepackage{booktabs}       
\usepackage{amsfonts}       
\usepackage{nicefrac}       
\usepackage{microtype}      
\usepackage{xcolor}         
\usepackage{hyperref}  
\usepackage{url}       
\usepackage{booktabs}     
\usepackage{amsfonts}   
\usepackage{amsmath}
\usepackage{nicefrac}      
\usepackage{microtype}
\usepackage{multirow}
\usepackage{wrapfig}
\usepackage{threeparttable}
\usepackage{graphicx}
\usepackage{xcolor} 
\usepackage{enumitem}
\usepackage{multicol}

\newcommand{\sref}[1]{Section.~\ref{#1}}

\newcommand{\figref}[1]{Fig.~\ref{#1}}
\newcommand{\figureref}[1]{Figure~\ref{#1}}

\newcommand{\tableref}[1]{Table~\ref{#1}}

\let\oldFootnote\footnote
\newcommand\nextToken\relax

\renewcommand\footnote[1]{%
    \oldFootnote{#1}\futurelet\nextToken\isFootnote}

\newcommand\isFootnote{%
    \ifx\footnote\nextToken\textsuperscript{,}\fi}

\title{Domain Generalization Strategy to Train Classifiers Robust to Spatial-Temporal Shift}

\author{%
   Minseok Seo\thanks{Equal contribution}\\
   SI Analytics\\
   \texttt{minseok.seo@si-analytics.ai}
   \And
   Doyi Kim\footnotemark[1]\\
   SI Analytics\\
   \texttt{doyi@ewhain.net}\\
   \And
   Seungheon Shin\\
   SI Analytics\\
   \texttt{shshin@si-analytics.ai}\\
   \And
   Eunbin Kim\\
   SI Analytics\\
   \texttt{ebkim@si-analytics.ai}\\
   \And
   Sewoong Ahn\\
   SI Analytics\\
   \texttt{anse3832@si-analytics.ai}\\
   \And
   Yeji Choi\thanks{Corresponding author}\\
   SI Analytics\\
   \texttt{yejichoi@si-analytics.ai}\\
}
\begin{document}

\maketitle

\begin{abstract}
Deep learning-based weather prediction models have advanced significantly in recent years.
However, data-driven models based on deep learning are difficult to apply to real-world applications because they are vulnerable to spatial-temporal shifts.
A weather prediction task is especially susceptible to spatial-temporal shifts when the model is overfitted to locality and seasonality.
In this paper, we propose a training strategy to make the weather prediction model robust to spatial-temporal shifts.
We first analyze the effect of hyperparameters and augmentations of the existing training strategy on the spatial-temporal shift robustness of the model.
\textcolor{black}{Next}, we propose an optimal combination of hyperparameters and augmentation based on the analysis results and a test-time augmentation.
We performed all experiments on the \textit{W4C'22 Transfer dataset} and achieved the 1st performance.
\end{abstract}
\section{Introduction}
Weather is a factor that affects the overall society, from transportation, energy, and agriculture to logistics and cultural industries.
Therefore, weather forecasting has become an essential technology in modern society.
Since Numerical Weather Prediction (NWP) models were first proposed in the 1920s~\cite{lynch2008origins}, the performance of NWP-based forecasting models has been steadily improving~\cite{haiden2018evaluation}.
Recently, extreme weather events have occurred frequently in every region across the globe ~\cite{masson2022regional}, so how to respond to climate-related risks has become an essential  topic of discussion. 
%
%
In particular, when meteorological phenomena occur locally, it is necessary to make predictions using regional-scale models with higher resolution than global-scale models.
However, regional models are costly because they require recalibration of parameters to apply to regions with different topography ~\cite{frnda2022ecmwf}.
%

%
%

%

In recent years, deep learning has been successfully applied to time series prediction using high-resolution images and has been found to take much less computing cost and time~\cite{pathak2022fourcastnet}.
Therefore, the deep learning-based approach is attracting attention as a key to solving the limitations of NWP models in weather forecasting.

Xingjian~\textit{et al.}~\cite{shi2015convolutional} pointed out the problem that the existing FC-LSTM did not capture spatial correlations adequately and proposed ConvLSTM to solve it.
ConvLSTM captured spatiotemporal correlations, showed significant results in weather forecasting, and received great attention.
Sønderby~\textit{et al.} suggested MetNet~\cite{sonderby2020metnet}, which receives a region larger than the target region as an input to design a model considering a large context. 
This architecture has become mainstream in deep learning-based weather forecasting.
Klocek~\textit{et al.} proposed MS-nowcasting~\cite{klocek2021ms} that combines the advantages of convLSTM and MetNet.
Despite these great advances, that models still have many challenging points to be resolved.

Unlike human vision systems~\cite{hendrycks2019benchmarking}, deep learning based data-driven models significantly degrade performance when different data is input from the data distribution in the training phase.
%
\textcolor{black}{In particular, weather forecast models, which have different weather and climate characteristics depending on the region, are sensitive to spatial-temporal changes and are challenging to apply in practice.}
%
In \textit{weather4cast 2022 competition}~\cite{10.1145/3459637.3482044}, the \textit{W4C'22 Transfer}~\cite{9672063} dataset is proposed to address these issues.
%
%
\textcolor{black}{The \textit{W4C'22 Transfer} training dataset consists of the regions R15, R34, R76, R4, R5, R6, and R7}
\textcolor{black}{The test dataset consists of those not included in the training dataset, which are R8, R9, and R10 regions}
%
\textcolor{black}{Thus, achieving high performance on the dataset must be robust to spatial-temporal data shifts.}
%

In this paper, we experiment and analyze the training strategy to make the deep learning-based weather forecast model robust against spatial-temporal shifts.
To improve domain generalization performance, existing deep learning-based recognition models mainly used data augmentation~\cite{hendrycks2019augmix} and ensemble methods.
However, since weather forecasting is based on the physical values of satellite and radar data, photometric augmentation cannot be applied, unlike RGB images.
In addition, since the capacity of weather data is very large, it is not easy to learn multiple models from scratch to ensemble differently from general recognition models.
We provide baselines on how to utilize photometric and geometric augmentation to solve these problems and propose a method to achieve the effect of an ensemble with a single model.

To sum up, our major contributions are as follows: 
\begin{itemize} 
\item We provide a basic selection policy for utilizing photometric and geometric augmentation for weather forecasting.

\item We propose a method that can produce the ensemble effect with a single deep learning-based weather forecasting model.

\item We find that existing deep learning-based weather forecast models have no spatial-temporal dependency, unlike reality, and propose a smooth loss that allows the model to have the spatial-temporal dependency.

\end{itemize}
\begin{figure*}[!t]
  \centering
  \includegraphics[width=1.0\linewidth]{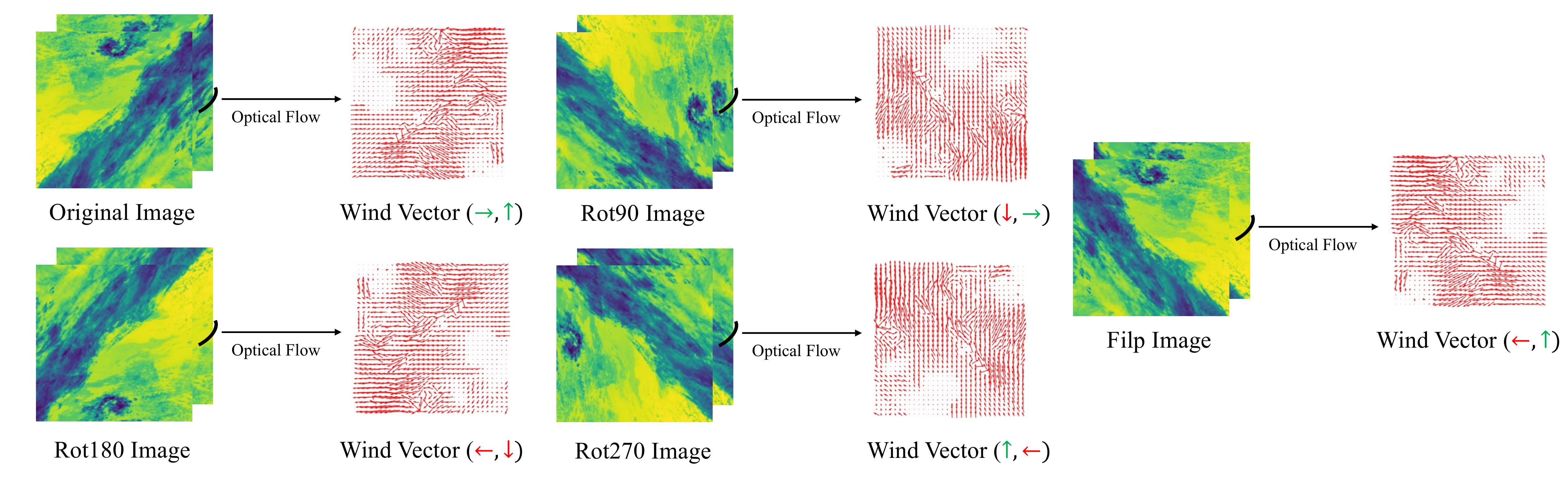}
  \caption{\textcolor{black}{A comparison of the results of applying the defined geometric augmentation policy. The direction of optical flow between $t_{0}$ and $t_{4}$ in the left panel of (a) is shown on the right. Figures (b) to (e) are the results of rotating and flipping the (a). The wind vector below each right panel represents the main wind direction, with the same direction as the original image indicated by a green arrow and the opposite direction by a red arrow.}}
  \label{fig:main}
\end{figure*}
\section{Method}
In this section, the data augmentation policy for making models robust to spatial-temporal shifts is described in detail in ~\sref{sec:aug}
The geometric augmentation ensemble, which has the ensemble effect of multiple models with a single model, is described in section ~\sref{sec:geo}.
%
%
\textcolor{black}{Finally, spatial-temporal smooth loss, which enables deep learning models to predict with spatial-temporal consistency, is described in detail in ~\sref{sec:gaus}}

\subsection{Augmentation Policy for Weather Forecasting}
\label{sec:aug}
Random color jittering, random crop, and random rotation augmentation are frequently used to prevent overfitting in general computer vision recognition tasks.
However, these methods remain \textcolor{black}{obscurity} without a well-defined or clear policy on how they work in physical fields~\cite{rasp2020weatherbench}.

From regional characteristics, we found clues of augmentation policy for weather forecasting.
\textcolor{black}{~\figureref{fig:main} shows the optical flow direction between $t_{0}$ and $t_{4}$ in the satellite image and compares the result of the geometric augmentation. Optical flow is the pattern of object motion between two consecutive images. As shown in the figure, if the original image((a)) is rotated $180^{\circ}$((c)), the moving direction of the cloud becomes the exact opposite of the original image. However, $90^{\circ}$, $270^{\circ}$ ((b) and (d)), and flip ((e)) overlapped the original image in at least one moving direction. Due to the regionality of test areas, it rarely moves in the opposite direction of the original image in the test set.}
%
%
%
Thus, training on out-of-distribution data, which cannot exist in the test set, hinders the convergence of the model.
We \textcolor{black}{decide to} use only augmentation policies that overlap at least one direction with the movement direction of the original image for geometric augmentation considering regionality.
The types of geometric augmentation that we used that \textcolor{black}{depend on} the regionality of Europe are as follows:

\begin{multicols}{3}
\begin{itemize}
    \item \textit{Rot90}
    \item \textit{Rot180}+\textit{Vertical flip}
    \item \textit{Rot270}
    \item \textit{Rot270}+\textit{Vertical flip}
    \item \textit{Vertical flip}
\end{itemize}
\end{multicols}

\subsection{Geometric Augmentation Ensemble}
\label{sec:geo}
\begin{figure*}[!t]
\label{figure:figure1}
  \centering
  \includegraphics[width=0.8\linewidth]{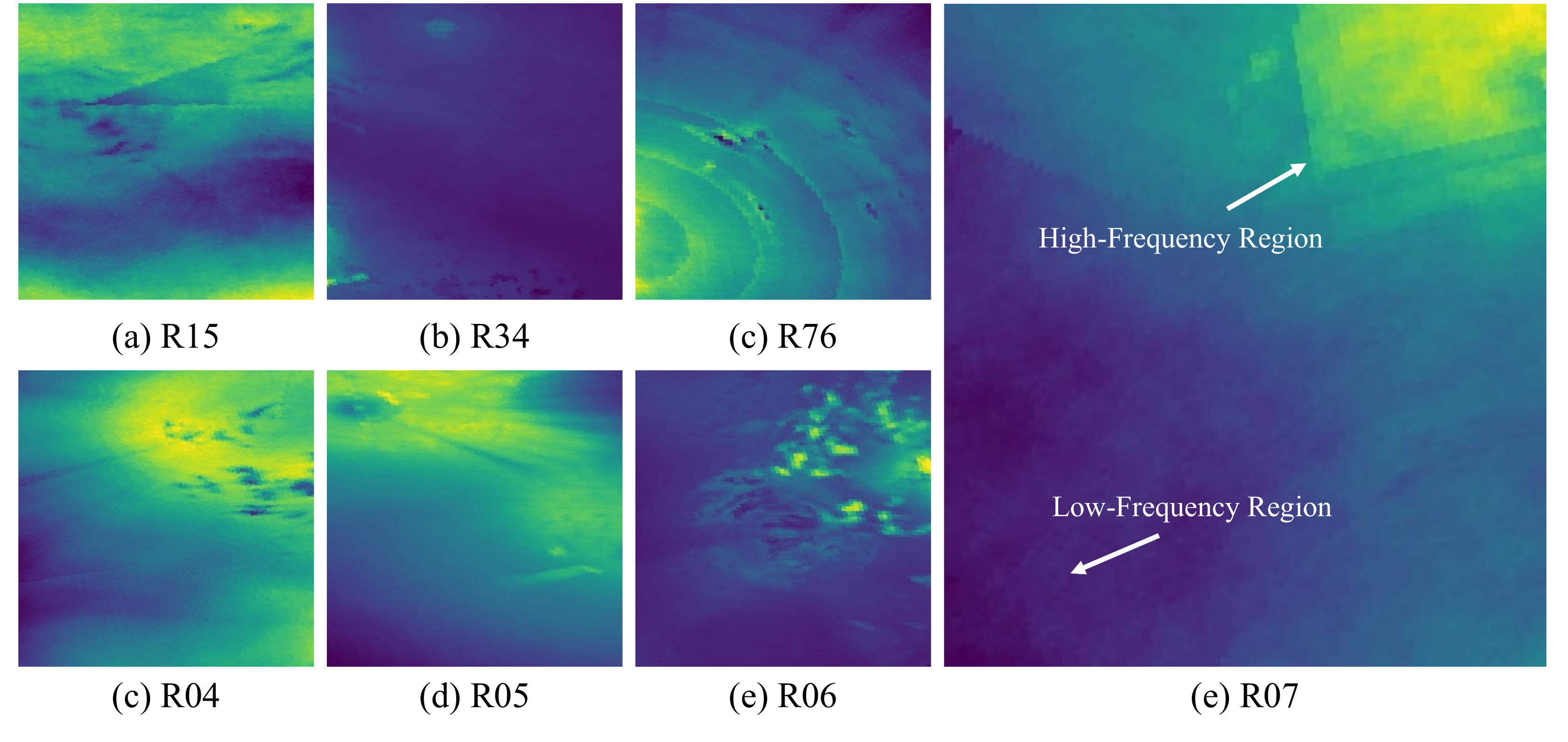}
  \caption{Quantitative comparison of spatial continuity in the training set of \textit{W4C'22 Transfer} 2019 dataset. In each region, the closer it is to green, the higher the frequency of rain.}
  \label{fig:main2}
\end{figure*}
Weather forecasting is affected by regionality. 
Using regionality (\textit{e.g.} lowering the threshold for areas with frequent rain and raising the threshold for areas with less rainfall) can improve the accuracy of weather forecasting, but overfitting  regionality makes it vulnerable to spatial shifts.
~\figureref{fig:main2} shows \textcolor{black}{the rain frequency for each region across the 2019 training dataset.}
As shown, there are areas with frequent and infrequent rain in each panel region.
\textcolor{black}{In particular, in the R07 and R34 regions, it can be seen that the influences of regionality appear relatively clearly.}

%
The geometric augmentation ensemble is a way to mitigate overfitting regionality predictions.
Given a weather forecast model $\mathcal{F}(.)$ trained with the geometric augmentation policy mentioned in ~\sref{sec:aug} and satellite data $x \in \mathbb{R}^{C \times T_{in} \times H \times W}$, the geometric augmentation ensemble is as follows:
\begin{equation}
  y_{ensemble} = \frac{1}{n}\sum_{i=1}^{n}{Aug^{-1}_{i}(\mathcal{F}(Aug_{i}(x)))}
  \label{eq:gae}
\end{equation}
where $Aug$ includes the augmentation policies mentioned in ~\sref{sec:aug} and no augmentation, and $Aug^{-1}$ is the inverse transformation of augmentation.

\subsection{Spatial-Temporal Smooth Loss}
\label{sec:gaus}
In the real-world rain process, if it rains at time $t_{0}$, there is a high probability that it will also rain at time $t_{1}$.
Conversely, if it rained at time $t_{1}$, it may have rained at time $t_{0}$, or there may have been prognostic symptoms.
Likewise, the closer the locations are spatial, the more likely they will be affected.
However, deep learning models cannot \textcolor{black}{correctly} reflect spatial-temporal dependence because they have poor calibration~\cite{guo2017calibration} performance unless explicit constraints are applied.
Basically, the object function of the semantic segmentation-based weather forecasting model is defined as follows:
\begin{equation}
  \mathcal{L}_{bce} = -\sum_{h,w}(y\log{\mathcal{F}(x)}+(1-y)\log(1-\mathcal{F}(x))
  \label{eq:overview5}
\end{equation}
where $y \in \mathbb{R}^{1 \times T_{out} \times \frac{H}{6} \times \frac{W}{6}}$ is a binary classification label. 
However, when training with the above objective function, $\mathcal{F}(.)$ is optimized to minimize cross entropy per pixel, but the spatial-temporal correlation is not considered.

The spatial smooth loss to consider spatial correlation is defined as follows:
\begin{equation}
  \mathcal{L}_{spatial} = \frac{1}{h \times w} \sum_{i,j}^{h,w}(||P_{(i,j)}-K_{3 \times 3}(P_{(i,j)})||)
  \label{eq:overview3}
\end{equation}
where $P$ is the probability value of applying sigmoid to the output of $\mathcal{F}(.)$, and $K$ is the sum in the square kernel.
Note that in all experiments in this paper, the kernel size was fixed at $3 \times 3$, but it is a hyperparameter that needs to be adjusted according to the spatial resolution of the output.

Temporal smooth loss considering temporal correlation is defined as below:
\begin{equation}
  \mathcal{L}_{temporal} = \frac{1}{T_{out}-1} \sum_{i=0}^{T_{out}-1}||P^{i}-P^{i+1}||
  \label{eq:smooth}
\end{equation}
Therefore, the final objective function of model $\mathcal{F}$ is defined as:
\begin{equation}
  \mathcal{L}_{total} = \mathcal{L}_{bce}+ \alpha\mathcal{L}_{spatial} +\beta\mathcal{L}_{temporal}
  \label{eq:smooth22}
\end{equation}
where $\alpha$ and $\beta$ are hyperparameters that control each element. In this paper, 0.1 was used for both factors.

\section{Experiments}
We conducted all experiments using the training set, validation set, and held-out set of the \textit{W4C’22 Transfer} dataset.
In this section, we experimentally verify our augmentation policy, geometric augmentation ensemble, and spatial-temporal smooth loss.
The performance evaluation metric of all experiments was selected as mean intersection over union (mIoU).
The code is available on \footnote{\tiny	{https://github.com/seominseok0429/W4C22-Simple-Baseline-for-Weather-Forecasting-Using-Spatiotemporal-Context-Aggregation-Network}}
\subsection{Experimental Setting}
\paragraph{\textit{W4C’22 Transfer} Training Datasets} consist of satellite data and radar ground truth data collected from 7 regions in Europe in 2019 and 2020.
The satellite data was obtained from a geostationary meteorological satellite operated by the European Organization for the Exploitation of Meteorological Satellites (EUMETSAT).
The radar data is obtained from the Operational Program for Exchange of Weather Radar Information (OPERA) radar network.
The satellite data consists of visible, infrared, and water vapor bands modalities (\textit{e.g.}, IR016, IR039, IR087, IR097, IR108, IR120, IR134, VIS006, VIS008, WV062, WV073).
%
Each spectral band provides different information depending on their absorption and scattering characteristics with atmosphere composition.
For instance, VIS channels represent information about albedo, shape, and roughness using solar radiation, but only available during the daytime.
IR channels \textcolor{black}{sense} re-emitted object radiation and can be converted to brightness temperature, and WV channels represent the amount of water vapor in the mid and upper atmosphere.
It is commonly used to detect clouds from the background and analyze cloud properties using single and multi-channel.
The satellite image covers a spatial resolution of 12 km x 12 km with 15-minute intervals.
The radar image consists of $252 \times 252$ pixels, the same as the satellite data, but covers only a central region of the satellite image with a spatial resolution of 2 km x 2 km (target region).

\paragraph{\textit{W4C’22 Transfer} Evaluation Datasets} consist of regions (spatial shift) R08, R09, R10 that are not included in the training set, and R15, R34, R76, R04, R05, R06, R07, R08, R09 and R10 in 2021 (temporal shift).

\paragraph{Implementation detail} We adopt the SImple baseline for weather forecasting using
spatiotemporal context Aggregation Network (SIANet)~\cite{seo2022simple} as the weather forecast model in all experiments.
To train SIANet, the batch size of one GPU was set to 16, and FP16 training was used.
In addition, 90 epochs were used and 4 positive weights of binary cross entropy were used.
The initial learning rate was set to 1e-4, weight decay to 0.1, and dropout rate to 0.4.
AdamW was used as the optimizer, and the learning rate was reduced by 0.9 when the loss was higher than the previous epoch validation loss.
All experiments were performed on Nvidia A100 $\times$ 8 GPUs.

\subsection{Main Results}
\begin{table}[h!]
\renewcommand{\arraystretch}{1.5}
\caption{\textit{W4C’22 Transfer} Held-out Leaderboard results.}
\label{tab:main}
\centering
\resizebox{0.99\columnwidth}{!}{%
\begin{tabular}{c|cccccccccccccccc|c}
\hline
\hline
\multicolumn{1}{l|}{} & \multicolumn{16}{c|}{\textbf{\textit{W4C'22 Transfer} Datasets}}                                                                                                          &      \\ \cline{2-17}
\textbf{Method}                & \multicolumn{3}{c|}{\textbf{2019}}               & \multicolumn{3}{c|}{\textbf{2020}}               & \multicolumn{10}{c|}{\textbf{2021}}                                          & \textbf{mIoU} \\ \cline{2-17}
\multicolumn{1}{l|}{} & R08  & R09  & \multicolumn{1}{c|}{R10}  & R08  & R09  & \multicolumn{1}{c|}{R10}  & R15  & R34  & R76  & R04  & R05  & R06  & R07  & R08  & R09  & R10  &      \\ \hline
SIANet                & \textbf{21.0} & \textbf{31.6} & \multicolumn{1}{c|}{\textbf{29.7}} & 27.7 & \textbf{20.1} & \multicolumn{1}{c|}{22.1} & 27.8 & 24.5 & 29.1 & \textbf{23.0} & 30.2 & 42.8 & \textbf{24.3} & \textbf{24.6} & 27.0 & \textbf{28.0} & \textbf{27.1} \\
meteoai               & 16.2 & 28.6 & \multicolumn{1}{c|}{28.9} & \textbf{29.3} & 19.5 & \multicolumn{1}{c|}{24.1} & \textbf{27.9} & 24.3 & 29.5 & 21.7 & \textbf{32.0} & \textbf{45.2} & 22.3 & 22.0 & \textbf{27.2} & 25.9 & 26.5 \\
FIT-CTU               & 20.4 & 29.5 & \multicolumn{1}{c|}{28.1} & 25.1 & 16.3 & \multicolumn{1}{c|}{24.0} & 27.0 & \textbf{25.0} & \textbf{29.7} & 21.0 & 30.4 & 42.8 & 22.4 & 21.3 & 23.3 & 23.5 & 25.6 \\
KAIST-CILAM           & 14.8 & 27.0 & \multicolumn{1}{c|}{29.0} & 25.9 & 19.2 & \multicolumn{1}{c|}{\textbf{27.7}} & 18.5 & 20.2 & 28.3 & 21.6 & 26.7 & 32.5 & 20.1 & 17.8 & 23.1 & 24.0 & 23.5 \\
Baseline           & 12.4 & 19.3 & \multicolumn{1}{c|}{21.1} & 22.9 & 14.4 & \multicolumn{1}{c|}{21.6} & 16.7 & 17.3 & 22.6 & 17.0 & 26.9 & 32.0 & 15.7 & 12.8 & 23.2 & 21.3 & 19.8 \\
\hline \hline
\end{tabular}%
}
\end{table}
\paragraph{Results} ~\tableref{tab:main} is the \textit{W4C’22 Transfer} held-out leaderboard. As shown in the table, SIANet achieved the highest performance in about 8 regions and achieved the highest performance with 27.1 mIoU.
Note that the SIANet in ~\tableref{tab:main} applies all of the augmentation policy, geometric augmentation ensemble (original, flip), and spatial-temporal smooth loss.

\paragraph{Discussion}
In the 2020 dataset, our proposed method showed lower performance improvement compared to the baseline.
We empirically found that the training loss differs by year and region.
In other words, the training difficulty is different for each year and region, and more weight is applied to regions with large losses.
In our future work, we plan to study loss regularization methods.
\subsection{Ablation Study}
\begin{table}[h!]
\centering
\begin{minipage}[l]{0.4\linewidth}
\centering
\caption{Component ablation study}
\label{tab:abl1}
\resizebox{1.0\columnwidth}{!}{%
\begin{tabular}{c|ccc|c|c}
\hline \hline
\textbf{Method}                      & \multicolumn{3}{c|}{\textbf{Componets}}                                      & \textbf{mIoU}                  & \textbf{Gain}                 \\ \cline{2-4}
                            & STL                   & AP                  & GAE                   &                       &                      \\ \hline
\multicolumn{1}{l|}{SIANet} & \multicolumn{1}{l}{} & \multicolumn{1}{l}{} & \multicolumn{1}{l|}{} & \multicolumn{1}{l|}{25.4} & \multicolumn{1}{l}{+0}      \\
\multicolumn{1}{l|}{SIANet} & \multicolumn{1}{l}{\checkmark} & \multicolumn{1}{l}{} & \multicolumn{1}{l|}{} & \multicolumn{1}{l|}{26.2} & \multicolumn{1}{l}{+0.8}      \\
\multicolumn{1}{l|}{SIANet} & \multicolumn{1}{l}{\checkmark} & \multicolumn{1}{l}{\checkmark} & \multicolumn{1}{l|}{} & \multicolumn{1}{l|}{27.0} & \multicolumn{1}{l}{+2.0} \\
\multicolumn{1}{l|}{SIANet} & \multicolumn{1}{l}{\checkmark} & \multicolumn{1}{l}{\checkmark} & \multicolumn{1}{l|}{\checkmark} & \multicolumn{1}{l|}{27.9} & \multicolumn{1}{l}{+2.9} \\ 
\hline
\hline
\end{tabular}}
\end{minipage}
\hspace{1.0cm}
\begin{minipage}[r]{0.4\linewidth}
\centering
\caption{Experimental results of random augmentation and inverse policy augmentation}
\label{tab:abl2}
\resizebox{\columnwidth}{!}{%
\begin{tabular}{l|ll|l}
\hline
\hline
\multicolumn{1}{c|}{\textbf{Method}} & \multicolumn{2}{c|}{\textbf{Componets}}                                  & \multicolumn{1}{c}{\textbf{mIoU}} \\ \cline{2-3}
\multicolumn{1}{c|}{}       & \multicolumn{1}{c}{Random} & \multicolumn{1}{c|}{Invers Policy} & \multicolumn{1}{c}{}     \\ \hline
SIANet                      &      \multicolumn{1}{c}{\checkmark}                      &                                    & 25.8                     \\
SIANet                      &                            &              \multicolumn{1}{c}{\checkmark}                      & 24.3                     \\ \hline \hline
\end{tabular}%
}
\end{minipage}

\end{table}
\paragraph{Results} ~\tableref{tab:abl1} is an ablation study of the effects of our proposed spatial-temporal smooth loss (STL), augmentation policy (AP), and geometric augmentation ensemble.
As a result of the experiment, it was experimentally shown that the performance of STL, AP, and GAE was improved and that there was a complementary relationship.
~\tableref{tab:abl2} shows the experimental results of random geometric augmentation and inverse policy augmentation. As shown in the table, the performance of the inverse policy is greatly reduced.
These experimental results indicate that augmentation does not improve the weather forecasting model performance.
Note that all experiments were performed on the validation set.
\paragraph{Discussion} We experimentally found that the accuracy tendency of regions for each year varies according to the $\alpha$ and $\beta$ weights of spatial-temporal smooth loss.
For example, the three groups of (R04, R05, R15), (R5, R10, R76), and (R6, R9, R34) regions had the same performance trend, but the different groups showed a trade-off relationship. 
As a result of analyzing the coordinates provided, (R04, R05, R15) and (R5, R10, R76) are adjacent to the sea on their left side, but (R6, R9, R34) are located in inland areas.
We could use the ensemble after training by setting different weights to improve performance. Still, it was not used in this challenge because it is not appropriate for the ensemble while looking at test performance in the domain generalization setting.
However, in our future work, we will analyze the spatial-temporal correlation according to regional characteristics and address the shortcomings.

\subsection{Qualitative results}
\begin{figure*}[!t]
  \centering
  \includegraphics[width=0.8\linewidth]{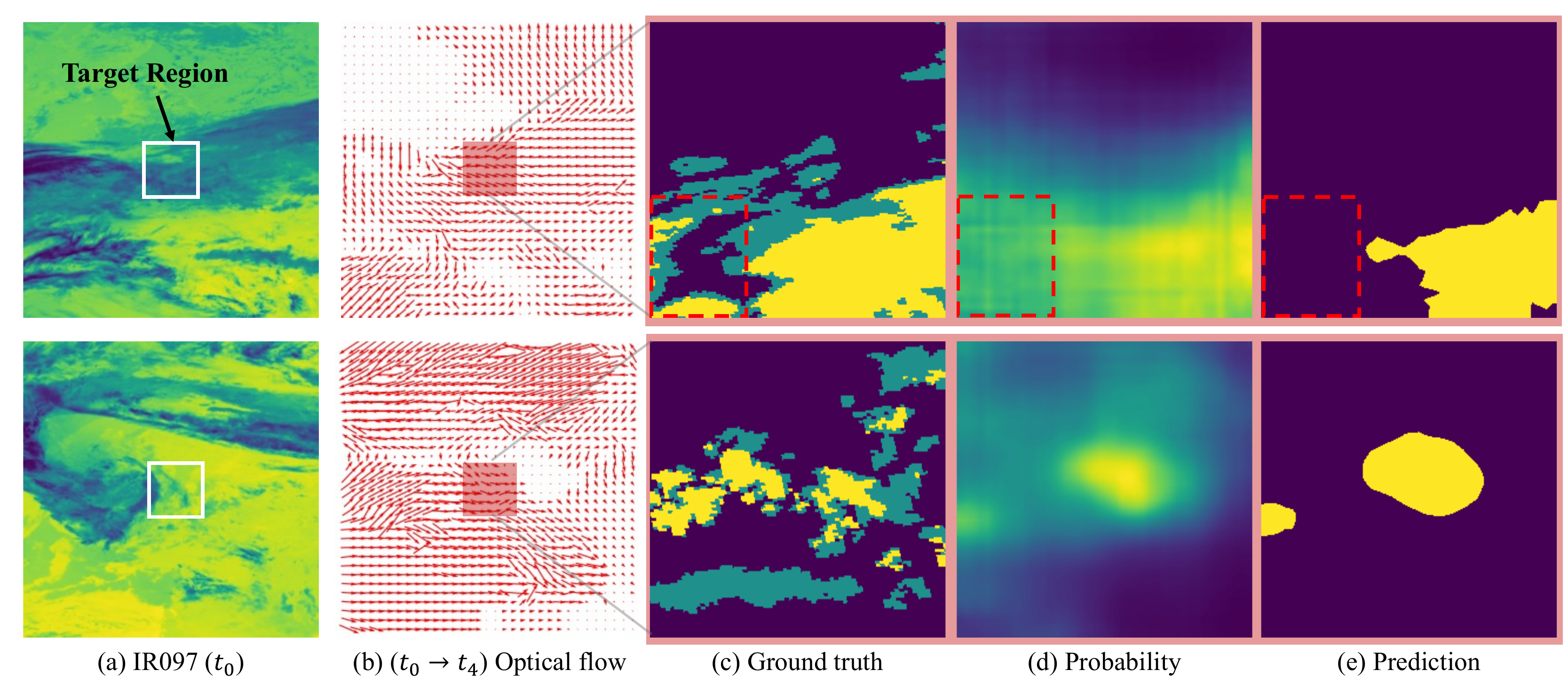}
  \caption{Quantitative test results.
  Note that we used all 11 channels in all experiments but only show IR097 due to space constraints.
  The optical flow indicates the direction of movement between the $t_{0}$ and the $t_{4}$. The yellow part of the ground truth is the area where the precipitation is 0.2 mm/hr or more, and the green part is 0.01 mm/hr or more. (a) and (b) have low resolution (12 km x 12 km), and (c), (d), and (e) have high resolution (2 km x 2 km).
}
  \label{fig:main3}
\end{figure*}
\paragraph{Results} ~\figureref{fig:main3} is the qualitative result of SIANet.
%
%
As shown in the ~\figref{fig:main3}-(d), SIANet detects the locations of 0.2mm/h rainfall (in yellow pixels) as high-confident regions. 
As we intended, the area around the rain area had a relatively high probability, even where it did not rain.
That is, the precipitation probability is shown as a continuous value rather than a discrete probability.
In addition, since most of the low-confident part of the probability map is the no-rain part of ground truth, it was shown that no-rain could be predicted.
\paragraph{Discussion} In a situation such as the red box in ~\figref{fig:main3}-(e), SIANet could not predict the presence or absence of rain in most cases.
If you zoom in on the part corresponding to the red box of IR097 and optical flow, the cloud moves from the upper left to the lower right. 
However, there seems to be no cloud in the upper left at $t_{0}$ that is likely to be a rainfall cloud.
Therefore, it is presumed that SIANet does not predict newly generated or developing rain clouds.
We will propose a model that considers these aspects in our future work.

\section{Conclusions}
In this paper, we proposed a strategy to train a model that is robust to spatial-temporal shifts in deep learning-based weather forecasting.
We pointed out that the existing augmentation policy was not \textcolor{black}{correctly established} in the field of weather forecasting.
Thus, we proposed the new augmentation policy and verified its effect experimentally.
In addition, it was shown that the weather forecasting model could be overfitted to topography through the distribution of the training dataset. To mitigate it, the geometric augmentation ensemble method was proposed.
Finally, we indicated the problem that the existing cross-entropy objective function could not capture the spatial-temporal relationship and suggested a spatial-temporal smooth loss.
These three training strategies achieved state-of-the-art on the \textit{W4R'22 Transfer} dataset.
We hope this training strategy will contribute to developing deep learning-based weather forecasting models.

{\small
\bibliographystyle{plain}
\bibliography{egbib}
}

\end{document}